\documentclass{article}

    \PassOptionsToPackage{numbers, compress}{natbib}

    \usepackage[preprint]{neurips_2022}

\usepackage[utf8]{inputenc} 
\usepackage[T1]{fontenc}    
\usepackage{hyperref}       
\usepackage{url}            
\usepackage{booktabs}       
\usepackage{amsfonts}       
\usepackage{nicefrac}       
\usepackage{microtype}      

\newcommand{\norm}[1]{\left\vert \left\vert #1 \right\vert \right\vert}

\newcommand{\rev}[1]{#1}
\newcommand{\revtwo}[1]{#1}
\usepackage{bm}
\usepackage{amsthm,amsmath}
\usepackage{tikz}
\usetikzlibrary{shapes,arrows,shadows}

\newtheorem{theorem}{Theorem}
\newtheorem{definition}[]{Definition}
\newtheorem{lemma}[]{Lemma}

\title{The robust way to stack and bag: the local Lipschitz way}

\author{%
  Thulasi Tholeti \qquad Sheetal Kalyani\\
  Department of Electrical Engineering\\
  IIT Madras\\
  \texttt{\{ee15d410, skalyani\}@ee.iitm.ac.in} \\
}

\begin{document}

\maketitle

\begin{abstract}
    Recent research has established that the local Lipschitz constant of a neural network directly influences its adversarial robustness. We exploit this relationship to construct an ensemble of neural networks which not only improves the accuracy, but also provides increased adversarial robustness. The local Lipschitz constants for two different ensemble methods - bagging and stacking - are derived and the architectures best suited for ensuring adversarial robustness are deduced. The proposed ensemble architectures are tested on MNIST and CIFAR-10 datasets in the presence of white-box attacks, FGSM and PGD. The proposed architecture is found to be more robust than a) a single network and b) traditional ensemble methods.
\end{abstract}

\section{Introduction}
Neural networks, especially deep learning, is known for its exceptional performance in solving problems in various fields. However, recent studies show that they are prone to adversarial attacks where a seemingly small, malicious change introduced in the input can cause the network to misclassify the data point \cite{szegedy2013intriguing}. This poses a high level of threat especially for critical fields such as self-driving or remote surgery. \rev{Adversarial attacks come in different forms based on various parameters such as availability of gradient to the attacker (white vs. black box attacks), norm of the input perturbation ($L_0$, $L_1$, $L_2$, $L_{\infty}$ norm attacks), number of steps (single step vs. iterative attacks) etc. (See \cite{akhtar2018threat,chakraborty2021survey} for a full survey).} Various techniques have been explored to ensure that neural networks are not susceptible to such attacks \cite{cisse2017parseval,tramer2018ensemble,ross2018improving,zhai2019adversarially,abbasi2020toward}. 

A promising solution was to perform adversarial training where adversarial examples were included in the training process \cite{goodfellow2014explaining}. However, the generation of these adversarial examples were computationally intensive. In \cite{ross2018improving}, the input of the gradient is regularized and this is shown to promote adversarial robustness. The authors of \cite{wu2021wider} explored if increasing the width of the network would correspond to higher robustness against adversaries. In \cite{yang2020closer}, the trade-off between accuracy and robustness was studied and it was observed that enforcing a local Lipschitz constraint on the network resulted in increased adversarial robustness. We discuss the relationship between adversarial robustness and Lipschitz constant in greater detail in the following section. 

\paragraph{Motivation} Constructing an ensemble network consisting of multiple base learners using methods such as stacking and bagging have been shown to be very useful in improving the performance of neural networks \cite{dong2020survey}. It is particularly useful when the base learners are diverse and learn different aspects of the problem. In this work, we explore the use of ensemble networks, specifically bagging and stacking, in the context of adversarial robustness. Typically, it is known that there is always a trade-off between robustness and accuracy \cite{yang2020closer}. As ensembles are known to improve accuracy, it is imperative that they are designed carefully to also ensure adversarial robustness. \revtwo{Here, we design a mathematical approach to constructing an ensemble of networks. Our approach allows us to choose the parameters of the ensemble, such as weights of the base learners for bagging and the meta-learner for stacking, so that the robustness of the ensemble is highly improved.} 

The use of ensemble of networks for adversarial robustness was introduced in \cite{abbasi2020toward}. \revtwo{This method requires the knowledge of the fooling matrix for specific adversaries. The fooling matrix quantifies the percentage with which a classifier is fooled to chose each of the other classes (except ground truth) under a particular adversarial attack. Then, for a $K$-class classification problem, $2K+1$ specialty networks are trained to classify specific subsets of classes correctly and a voting mechanism is proposed. In our work, we propose a way to ensemble existing base learners to achieve improved robustness over a single network as well as ensembles constructed without analysing the choice of parameters for ensembling. We differ from \cite{abbasi2020toward} in these major regards: 1) Our work does not require the knowledge of the fooling matrix or the exact adversarial attack, 2) It can be used to construct an ensemble of any number of pre-trained networks and 3) The choice of our ensemble parameters are derived based on local Lipschitz constant of the networks which are shown to have a direct influence over adversarial robustness.
}

\paragraph{Contributions}
For two of the popular ensemble methods - stacking and bagging, we derive the local Lipschitz constant as a function of the individual network constants. We further analyse the choice of weights for the weighted average and meta-learner in the case of bagging and stacking respectively. The performance of the analytical choice is then verified with simulations on the MNIST and CIFAR-10 datasets for selected adversarial attacks. \rev{For our simulations, we use FGSM and PGD attacks which are white-box attacks, i.e., adversary possess information regarding the target model weights and gradients.} We observe that the ensemble network designed by the proposed strategy displays better adversarial robustness while achieving better accuracy on clean samples. \revtwo{We also demonstrate the choice of parameters according to our analysis is crucial and robustness is affected if the parameters do not conform to the derived conditions.}

\section{Lipschitz constant and  adversarial robustness}
 Lipschitz constant is a parameter of utmost interest in quantifying robustness of a network as evidenced by the recent surge in literature \cite{yang2020closer, anil2019sorting, jordan2020exactly}. In this section, the evolution of the principal ideas concerning adversarial robustness and Lipschitz constant are explored. We begin with providing the formal definition of the Lipschitz constant.
\begin{definition}
        A function \(f:\mathbb{R}^m \rightarrow \mathbb{R}^n\)  is said to be \(L_f\)-Lipschitz continuous if  \(\forall \bm{x_1},\bm{x_2} \in \mathbb{R}^m\)
        \begin{equation} \label{eqn:LipDefn}
            \norm{ f(\bm{x_1}) -  f(\bm{x_2})} \leq L_f \norm{\bm{x_1-x_2}},
        \end{equation}
        and \(L_f\) is known as the Lipschitz Constant.
\end{definition}

The authors of \cite{szegedy2013intriguing} were one of the first to analyze instability of the network; it was suggested that the Lipschitz constant can be studied to determine the extent to which a perturbation at the input can affect the output. 
If the input-output relation of a \(K\)-layered network is given by \(\phi:\mathbb{R}^m \rightarrow \mathbb{R}^n\), \cite{szegedy2013intriguing} states that the network satisfies
\begin{equation} \label{eqn:szegedy}
    \norm{\phi(\bm x+ \bm a)-\phi(\bm x)} \leq \left(\prod_{i=1}^K L_i \right) \norm{\bm a}, \quad \bm x, \bm a \in \mathbb{R}^m 
\end{equation}
where $L_i$ refers to the Lipschitz constant of the $i_{th}$ layer. As the authors compute the upper bound of the Lipschitz constant and not the exact value, they emphasize that large bounds do not necessarily translate to inefficiency against adversarial samples whereas small bounds do guarantee adversarial robustness. However, the bounds determined by the product are very loose.

There have been many attempts to determine the exact Lipschitz constant of neural networks, not necessarily in the context of adversarial robustness. In \cite{scaman2018}, the authors show that the exact computation of Lipschitz constant of a neural network is NP-hard and propose a method to computationally determine the Lipschitz constant using automatic differentiation. This method is shown to provide very loose estimates for the upper bound for networks of higher dimensions. Alternatively, \cite{Fazlyab2019LipSDP} formulates the problem of estimating the Lipschitz constant as a semi-definite program, which should be solved mathematically to obtain the global minimum.

There has been interest in providing a guarantee based on Lipschitz constant which is better than the bound proposed in (\ref{eqn:szegedy}). For classification networks, the authors of \cite{weng2018evaluating} provide a theoretical guarantee using the Lipschitz constant of the difference of two output nodes using extreme value theory. In \cite{hein2017formal}, a lower bound on the magnitude of input perturbation is derived so that the resultant class remains unchanged. However, considering the Lipschitz constant of the function computed for its entire domain as an indicator for robustness while dealing with specific input was found to be restrictive. The authors preferred considering the local Lipschitz constant instead, which is formally defined below.

\begin{definition} \label{defn:llc}
A function \(f:\mathbb{R}^m \rightarrow \mathbb{R}^n\)  is \(\ell_f\)- locally Lipschitz continuous at radius $r$ if for each \(i =1,\cdots n\)
        \begin{equation} \label{eqn:lLipschitz constantDefn}
            \norm{ f(\bm{x_1}) -  f(\bm{x_2})} \leq \ell_f \norm{\bm{x_1-x_2}},
        \end{equation}
        for all $dist(\bm x_1,\bm x_2)\leq r$ and \(\ell_f\) is known as the local Lipschitz Constant.
\end{definition}
The observation that local Lipschitz constant is more indicative of adversarial robustness than the regular Lipschitz constant is also corroborated by \cite{wu2021wider}; the authors show that local Lipschitz constant is directly linked to perturbation stability which in turn relates to robustness. The authors of \cite{yang2020closer} show that although typically practical systems trade-off robustness and accuracy, it is possible to achieve both simultaneously. They also note that this can be achieved by imposing local Lipschitz conditions on the network function. In the next section, we take a closer look at how we can achieve both robustness and accuracy when constructing an ensemble of neural networks.

\section{Provable adversarial robustness for ensemble networks} \label{sec:sec3}
Among the vastly varied methods proposed to enhance the performance of neural networks, using an ensemble of networks and combining their decision to achieve a final output has shown promise in multiple fields \cite{ren2015ensemble,zhai2019adversarially,ardabili2019systematic}. Ensemble networks offer advantages from various vantage points \cite{ganaie2021ensemble}. Some methods (like bagging) focus on decreasing the variance of the prediction whereas others aid by lowering the bias. Ensemble networks also allow for the use of relatively shallow networks to match the performance of a single deep network \cite{veit2016residual}. 

In addition to all the above benefits, \cite{abbasi2020toward} recently showed that the use of ensemble networks with sufficient diversity also resulted in adversarial robustness. The final decision is arrived at through a voting mechanism. Alternatively, in our work, we choose the parameters of the ensemble (parameters used to combine base learners) based on the desired local Lipschitz constant of the network. To achieve this, we initially derive the local Lipschitz constant of an ensemble of networks. Further, we explore how to construct such ensembles to provide adversarial robustness. We focus on two of the popular ensemble methods: stacking and bagging.

\subsection{Lipschitz constant of a stacked network} \label{sec:sec3.1}
Stacking is a popular ensemble method which uses a meta-learner to process the outputs from base learners. It enhance the performance of the network by decreasing the bias in the network \cite{leblanc1996combining}. It is especially useful when the base learners learn different patterns from the data. Let us consider an ensemble of networks obtained by stacking \(n\) neural networks known as base networks. The outputs of the stacked networks, denoted by \(\bm y_i = f_i(\bm x) \in \mathbb{R}^k, i=1\cdots n\), are fed into a meta-learner which then produces the final output \(\bm z = g(\bm \vec y) \in \mathbb{R}^k\). Here,  \(\bm \vec y = [\bm y_1; \cdots \bm y_n]\) is a vertical stack of the output vectors from the neural networks. Note that all the neural networks are provided with the same input \(\bm x \in \mathbb{R}^d\). The architecture is illustrated in Fig. \ref{fig:stacked}.


\tikzset{%
  every neuron/.style={
    rectangle,
    draw,
    minimum height=1cm, 
    minimum width=2cm
  },
  neuron missing/.style={
    draw=none, 
    scale=4,
    text height=0.333cm,
    execute at begin node=\color{black}$\vdots$
  },
}
\begin{figure}[ht]
\centering
\begin{tikzpicture}[scale = 0.7,x=2cm, y=2cm, >=stealth]

\foreach \m/\l [count=\y] in {1,2,missing,3}
  \node [every neuron/.try, neuron \m/.try] (input-\m) at (0,2.5-\y) {};

\foreach \l [count=\i] in {1,2,n}
  \draw [<-] (input-\i) -- ++(-2,0)
    node [above, midway] {$\bm x$};

\foreach \l [count=\i] in {1,2,n}
  \draw [->] (input-\i) -- ++(2,0)
    node [above, midway] {$\bm y_\l$};

\draw (4,-2) rectangle (2,2.2) node[pos=.5] {Meta-learner};

\draw [->] (4,0) -- (5,0) node [above, midway] {$\bm z$};
    
    


\end{tikzpicture}
   \caption{Stacked ensemble of neural networks with a meta-learner}
    \label{fig:stacked}
\end{figure}
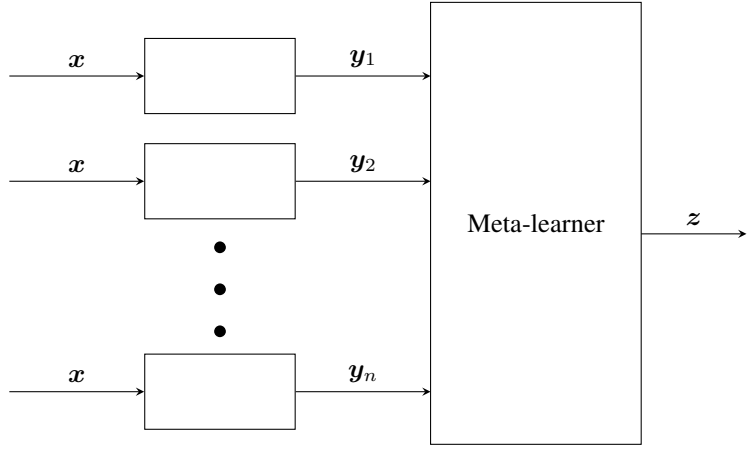

\paragraph{Assumptions} It is assumed that the neural networks as well as the meta learner are Lipschitz continuous and their Lipschitz constants are \(L_{f_i}, i = 1,\cdots, n\) and \(L_g\) respectively. Note that they can refer to both global and local Lipschitz constants based on the inputs considered. Here, we derive for a general framework.

\begin{theorem} \label{thm:stacking}
    Consider an ensemble of \(n\) neural networks each with a Lipschitz constant of \(L_{f_i}, i = 1,\cdots, n\) stacked with a meta-learner with a Lipschitz constant of \(L_g\). The Lipschitz constant of the architecture is \(L_g \sqrt{\sum_i L_{f_i}^2}\). 
\end{theorem}
\begin{proof}
We derive the Lipschitz constant of the ensemble network by considering two outputs of the network \(\bm z_1\) and \(\bm z_2\) corresponding to the inputs \(\bm x_1\) and \(\bm x_2\) respectively.
    \begin{align}
        \norm{\bm z_1 - \bm z_2} &= \norm{g(\bm \vec y_1) - g(\bm \vec y_2)}\\
        &\leq L_g\norm{\bm \vec y_1 - \bm \vec y_2}\\
         & = L_g \norm{\begin{bmatrix} f_1(\bm x_1) - f_1(\bm x_2)\\
        f_2(\bm x_1) - f_2(\bm x_2)\\
        \vdots
        f_n(\bm x_1) - f_n(\bm x_2)\\
        \end{bmatrix}}=L_g \sqrt{\sum_{i=1}^n (f_i(\bm x_1) - f_i(\bm x_2)})^2\\
        &\leq L_g  \sqrt{\sum_i L_{f_i}^2} \norm{\bm x_1 -\bm x_2}.
    \end{align}
    The final inequality is due to the Lipschitz continuity of the functions \(f_i\). This proves Theorem \ref{thm:stacking}. It is noteworthy that this bound is tight if the bounds of the individual Lipschitz constants are tight.
\end{proof}

\subsubsection{How to choose the meta-learner?}
To design the meta-learner such that provable adversarial robustness guarantees can be provided for an ensemble network, the Lipschitz constant of the ensemble should be lower than that of an individual network. In case we only have access to the upper bounds on the Lipschitz constants,  the upper bound on the ensemble should be lesser than the upper bound for any individual network. First, let us consider all base networks to be similar and hence, having the same Lipschitz constant \(L_f\). Using the result from Theorem \ref{thm:stacking}, 
\(
    \sqrt{n} L_g L_f \leq L_f \Rightarrow L_g \leq \frac{1}{\sqrt{n}}.\)
Now, let us consider the case where the networks are different. As the ensemble should have a lower Lipschitz constant when compared to the network with the minimum Lipschitz constant, 
\begin{equation} \label{eqn:stack_condition}
    L_g  \sqrt{\sum_i L_{f_i}^2} \leq L_{min} \Rightarrow L_g \leq \dfrac{L_{min}}{\sqrt{\sum_i L_{f_i}^2}}.
\end{equation}
Constructing such a meta-learner will ensure that the ensemble networks offers better robustness guarantees than an individual network.

\subsection{Lipschitz constant of a bagged network}
Bagging or model averaging uses an ensemble of networks that offer diversity. The diversity may be due to varied initialization techniques, hyperparameter tuning strategies, training data, etc. The outputs from the different models are averaged to minimize the model variance and offer better generalization. Note that the averaging may be simple averaging or weighted average. 
Let the outputs of the $n$ constituent base networks in the ensemble be denoted as \(\bm y_i = f_i(\bm x) \in \mathbb{R}^k, i=1\cdots n\); the output after model averaging is \(\bm z = \sum_{i=1}^n w_i \bm y_i\), where \(w_i\)'s are the weights assigned to each network.
\begin{theorem} \label{thm:bagging}
    Consider an ensemble of \(n\) neural networks each with a Lipschitz constant of \(L_{f_i}, i = 1,\cdots, n\) bagged with weights $w_i$. The Lipschitz constant of the architecture is \(L_z \leq \sum_{i=1}^n w_i L_{f_i}\). 
\end{theorem}
\begin{proof}
 The Lipschitz constant of the output \(L_z\) is
\begin{align}
    \norm{\bm z_1 - \bm z_2} &= \norm{\sum_{i=1}^n w_i \bm y_{1_i} - \sum_{i=1}^n w_i \bm y_{2_i}}
    = \norm{ \sum_{i=1}^n w_i (\bm y_{1_i} - \bm y_{2_i})}\\
    &\leq \sum_{i=1}^n  \norm{  w_i (\bm y_{1_i} - \bm y_{2_i})} = \sum_{i=1}^n w_i \norm{\bm y_{1_i} - \bm y_{2_i}} \leq \sum_{i=1}^n w_i L_{f_i} \norm{\bm x_1 - \bm x_2}
\end{align}
The triangle inequality and the fact that the weights are positive are used to derive the result. The Lipschitz constant of the bagged ensemble is derived as
\(
    L_z \leq \sum_{i=1}^n w_i L_{f_i}.
\)
\end{proof}

\subsubsection{How to choose weights?}
Ideally, the ensemble network should have a lesser Lipschitz constant than a single network to ensure greater robustness. Let us assume that all constituent networks have the same Lipschitz constant, \(L_f\) and simple averaging is performed (i.e., \(w_i = 1/n \forall i\)). Then the Lipschitz constant of the ensemble becomes:
\begin{equation}
    L_z = \dfrac{n L_f}{n} = L_f,
\end{equation}
which is the same as using a single network. This is intuitive as model averaging with the same network will result in the same output. Now, let us assume that the networks have different Lipschitz constants. Recall that lower local Lipschitz constant is indicative of better adversarial robustness. We can make the following statements: 1) Lipschitz constant of the ensemble can (at best) only be the minimum of the Lipschitz constants of the constituent networks and 2) to lower the Lipschitz constant of the ensemble, it should be preferable to assign higher weights to networks with lower Lipschitz constants and vice versa. We formalise this statement as follows.

\begin{lemma} \label{lemma:major}
Let the Lipschitz constants of the $n$ base networks be ordered as $\ell_1 \leq \ell_2 \cdots\leq \ell_n$ and denoted as vector $\bm \ell$. Consider two possible weight vectors $\bm w, \bm w' \in \mathbb{R}^n$ such that $\bm w \succ \bm w'$. Then, the Lipschitz constant of the bagged ensemble, $\bm \ell^T \bm w \geq \bm \ell^T \bm w'$.
\end{lemma}

\begin{proof}
\rev{\revtwo{The proof is based on the result in \cite[Lemma V]{raj2017aggregating} where it was given in the context of analysis of the HEDGE algorithm.} Let the components of vectors $\bm w$ and $\bm w'$ be arranged in decreasing order. By the definition of majorization \cite{marshall1979inequalities}, we say that $\bm w \succ \bm w'$ when $\sum_{i=1}^n w_i = \sum_{i=1}^n w'_i$ and $\sum_{i=1}^k w_i \geq \sum_{i=1}^k w_i, k=1,\cdots,n$. Let $m_{j}$ be some non-negative numbers. Then consider the sequence of inequalities,
$$
\begin{aligned}
m_{1} w_1 & \geq m_{1} \cdot w'_1 \\
m_{2} \cdot\left(w_1 +w_2 \right) & \geq m_{2} \cdot\left(w'_1+w'_2\right) \\
\vdots & \\
m_{K-1}\left(\sum_{k=1}^{K-1} w_k \right) & \geq m_{K-1} \cdot\left(\sum_{k=1}^{K-1} w'_k\right) \\
m_{K}\left(\sum_{k=1}^{K} w_k\right) &=m_{K} \cdot\left(\sum_{k=1}^{K} w'_k\right)
\end{aligned}
$$
Summing over all the terms,
$$
\sum_{k=1}^{K}\left(\sum_{j=k}^{K} m_{j}\right) \cdot w_k \geq \sum_{k=1}^{K}\left(\sum_{j=k}^{K} m_{j}\right) \cdot w'_k.
$$
Consider $\sum_{j=k}^{K} m_{j}=\ell_k$ and observe that $\sum_{j=k}^{K} m_{j} \geq \sum_{j=k+1}^{K} m_{j}$, we have $\sum_{k=1}^{K} \ell_k \cdot w_k \geq \sum_{k=1}^{K} \ell_k \cdot w'_k$.}
\end{proof}

This result shows that for a bagged ensemble to be effective, the weights should be spread out so that the highest weight corresponds to the lowest Lipschitz constant. It can be trivially extended to state that, for maximum adversarial robustness, the weight corresponding to the network with the least local Lipschitz constant should be made 1 while forcing the rest to 0. We note that although this boundary case reverts to the use of a single network, it is only optimal from the adversarial robustness, and does not provide other ensemble benefits. The result is particularly useful when designing a network with a target local Lipschitz constant.

\section{Simulation Results}
In this section, we evaluate the local Lipschitz constant, the adversarial accuracy and the clean test accuracy for different base networks and the ensembles. Unlike evaluating global Lipschitz function, which depends entirely on the function, local Lipschitz constant also depends on the input as well as the radius around the input as mentioned in Definition \ref{defn:llc}. Therefore, the local Lipschitz constants are reported as an empirical approximation computed using the inputs from the data sets. The empirical formulation for computing the local Lipschitz constant is given by the following quantity \cite{yang2020closer}
\begin{equation}
    \dfrac{1}{n} \sum_{i=1}^n \max_{\bm x_i' \in B_{\infty}(\bm x_i, \epsilon)} \dfrac{\norm{f(\bm x_i) - f(\bm x_i')}_1}{\norm{\bm x_i - \bm x_i'}_{\infty}}.
\end{equation}
Here, $B_{\infty}(\bm x_i, \epsilon)$ denotes a infinity-norm ball with center $x_i$ and radius $\epsilon$. 
\subsection{Adversarial attacks}
In our work, we focus on white-box attacks, i.e., attacks in which the adversary can access full information regarding the target model such as weights and gradients. White-box attacks are more potent when compared to black-box attacks where the adversary can only access the outputs. The Fast Gradient Sign Method (FGSM) is a white-box attack that produces a linear approximation of the loss function around the input \cite{goodfellow2014explaining}. For a data point $(\bm x, y)$ input into a function $f$ and with loss function $J$, the adversarial output
\begin{equation}
    \bm x^{adv} = \bm x + \epsilon \quad sign(\Delta J(f(\bm x), y))
\end{equation}
where $\epsilon$ is a parameter defined by the adversary. The FGSM is a single-step attack where the input is perturbed just once. An example of a stronger attack is the Projected Gradient Descent (PGD) attack which iteratively applies the following update, as presented in \cite{kurakin2016adversarial}
\begin{equation}
    \bm x^{adv}_{k+1} = Proj_{B_{\epsilon} (\bm x)}\left( \bm x^{adv}_{k} + \eta sign\left( \Delta_{x^{adv}_{k}}J(f(x^{adv}_{k}),y)\right)\right)
\end{equation}
where $Proj_{B_{\epsilon}( \bm x^{adv})} = \arg \min_{\bm x' \in B_{\epsilon}(\bm x)} \norm{x^{adv}-x'}_p$. Parameters such as number of iterations, $\epsilon$ and the step size $\eta$ are determined by the adversary. For simulating our adversarial attacks, we have used an $\epsilon$ of 0.1 for FGSM and 0.01 for PGD.

\paragraph{Simulation setup} We consider 3 base learners for each of the ensemble methods consisting of 2, 4 and 5 layers for feed forward networks (FNN) and convolutional neural networks (CNN) for the MNIST dataset. For the CIFAR-10 dataset, our base learners are constructed so that they have 3, 5, and 6 layers where the final two layers are feed-forward. Max pooling operation is performed after every convolutional layer. All the network layers employ ReLU activation. The base networks are trained using the Adam optimizer with the default learning rate and a batch size of 128 for 100 epochs. The categorical accuracy is employed as the metric for evaluation. Note that we do not claim that the proposed base learners are optimal or state-of-the-art. We consider them for their variation in local Lipschitz constants so that we can demonstrate our technique for ensembling base learners.

The meta learner in the stacked network is chosen so that the Lipschitz condition in \ref{eqn:stack_condition} is satisfied. The weights of the bagged network are formulated to be inversely proportional to their individual empirical local Lipschitz constant and are normalized so that they sum to 1. This choice of weights ensures that a higher weight is allotted to the base learner with the lowest local Lipschitz constant while also ensuring that the weights are sufficiently spread apart.

\begin{table}[h]
    \centering
        \caption{Local Lipschitz constant and test accuracy percentage for MNIST dataset}
    \label{tab:MNIST}
    \begin{tabular}{lllllllll}
        \hline
         & \multicolumn{4}{c}{Feed forward networks} &  \multicolumn{4}{c}{Convolutional neural networks}
         \\
         \hline
          &    &   \multicolumn{3}{c}{Accuracy Percentage}    &  & \multicolumn{3}{c}{Accuracy Percentage} \\
         \cmidrule(r){3-5} \cmidrule(r){7-9}
        Network &  LLC & Clean & FGSM & PGD &  LLC & Clean & FGSM & PGD\\
        \hline
        2 layer &  24.23 & 98.01 & 38.1 & 12.8 & 13.475	& 98.09 &	31.5 & 28.8
\\
        4 layer &  18.59 & 97.79 & 51.7 & 45.7& 4.647 &	98.72&	70.2 & 79.5
\\
        5 layer &  12.45 & 97.88 & 51.6 & 48.35&2.481&	98.91&	81.5&87.4
 \\
        \hline
        Bagged (proposed) &  14.21 & 98.21 & 58.7 & 52.9& 3.872	&98.93&	80.4& 86.47
\\
        Bagged (equal) &18.28 & 98.71 & 58.4 & 50.26& 6.399&	99.01	&70.9& 78.8
\\
        Bagged (reverse) &20.37 &  98.24 & 58.4 & 51.85& 7.197&	98.53	&41.4& 38.29
\\
        \hline
        Stacked (proposed) &  12.41 &    98.54  & 62.34 &   56.2 & 3.271 & 99.05& 90.28&  94.52\\
        Stacked (reverse) & 18.94 & 98.43 & 55.39 & 48.68&  9. 74 & 98.79 & 83.52& 79.31\\
        \hline
    \end{tabular}
\end{table}

\begin{table}[ht]
    \centering
        \caption{Local Lipschitz constant and test accuracy percentage for CIFAR-10 dataset using CNN}
    \label{tab:CIFAR}
    \begin{tabular}{lllll}
        \hline
          &   &   \multicolumn{3}{l}{Accuracy Percentage} \\
         \cmidrule(r){3-5} 
        Network & LLC & Clean & FGSM & PGD\\
        \hline
        3 layer & 16.85	& 58.02&	18.4&11.19\\	
        5 layer & 18.63	&68.64	&18.1	& 12.21
\\
        6 layer & 18.46	&72.41	&16.9	& 12.48
\\
        \hline
        Bagged (proposed) & 8.31  &73.18 & 17.4& 13.5 \\
        Bagged (equal) & 8.32  & 72.94&	17.4
 & 13.5\\
        Bagged (reverse)  &  8.34 & 72.91&17.3 & 13.52\\
        \hline
        Stacked (proposed) &  7.15 & 73.21 &19.31& 13.78\\
        Stacked (reverse) & 12.49 & 72.13& 17.16& 12.67 \\
        \hline
    \end{tabular}
\end{table}

\rev{To further demonstrate the utility of our work, we report the accuracy for ensemble networks when the proposed architecture for the ensemble networks are not chosen. Consider the case of bagging where our work suggests that the weights assigned to the base learners should be inversely proportional to the local Lipschitz constant of the network. We also demonstrate that in case we choose the weights differently, i.e., either opt for equal weights for the various base learners or reverse the proposed order, adversarial robustness of the resultant ensemble network is affected. Similarly, for the case of stacking, when the meta learner is chosen such that the condition in (\ref{eqn:stack_condition}) is not met, the adversarial performance of the ensemble decreases. This illustrates the importance of the choice of parameters while designing the network.} The accuracy reported on clean samples and samples under adversarial attack are tabulated, along with their respective empirical local Lipschitz constants in Tables \ref{tab:MNIST} and \ref{tab:CIFAR}.

\rev{From Table \ref{tab:MNIST}, we see that the accuracy improves while using an ensemble when compared to any one of the individual networks, as expected. The maximum performance is achieved when using the proposed choices for weights and the meta learner. Although the use of different parameters do yield an improvement in performance, we note that higher accuracy can be achieved with the proper choice. This is especially pronounced for the stacked network.} In Table \ref{tab:CIFAR}, the local Lipschitz constants of the base learners are similar. Although the values for stacked and bagged networks are similar, we can observe that the choice of parameters according to the proposed method almost always yields better results.

In both MNIST and CIFAR-10 datasets, we make the following observations
\begin{itemize}
    \item The accuracy for both FGSM as well as PGD attacks for the ensemble networks (both bagging and stacking) are significantly higher than any of the base learners.
    \item Stacking allows us to achieve a lower local Lipschitz constant when compared to any of the base learners whereas bagging only allows us to obtain a weighted average of the existing base learners.
    \item The choice of weights in case of bagging and the choice of the meta learner in stacking performed according to our analysis plays a significant role in ensuring adversarial robustness and provides intuition for the choice.
\end{itemize}

\section{Discussions} \label{sec:discussions}
In this work, we study how to exploit the relationship between local Lipschitz constant and adversarial robustness while designing ensemble networks. In particular, we derive the Lipschitz constant of an ensemble networks in terms of the Lipschitz constant of their constituents. This, in turn, led us to analysing and proposing the choice for weights and meta-learner for bagging and stacking. Our simulations show that the proposed choices result in increased robustness while facing white-box adversarial attacks such as FGSM and PGD while still reporting high accuracy on clean samples. Our work is currently limited to the use of pre-trained networks in the ensemble. Future directions to explore include performing a similar analysis for the choice of ensemble parameters while performing adversarial training. It would also be interesting to extend this analysis to other deep ensemble methods such as negative correlation based deep ensemble networks and decision fusion strategies. 

\bibliographystyle{IEEEtran}
\bibliography{ref.bib}

\end{document}